# Short-term probabilistic photovoltaic power forecast based on deep convolutional long short-term memory network and kernel density estimation

Mingliang Bai[1], Xinyu Zhao[2], Zhenhua Long[2], Jinfu Liu[2]*, Daren Yu[1,2]

1. Department of Control Science and Engineering, Harbin Institute of Technology, 150001, Harbin, Heilongjiang, China
2. School of Energy Science and Engineering, Harbin Institute of Technology, 150001, Harbin, Heilongjiang, China

**Abstract:** Solar energy is a clean and renewable energy. Photovoltaic (PV) power is an important way to utilize solar energy. Accurate PV power forecast is crucial to the large-scale application of PV power and the stability of electricity grid. This paper proposes a novel method for short-term photovoltaic power forecast using deep convolutional long short-term memory (ConvLSTM) network and kernel density estimation (KDE). In the proposed method, ConvLSTM is used to forecast the future photovoltaic power and KDE is used for estimating the joint probabilistic density function and giving the probabilistic confidence interval. Experiments in an actual photovoltaic power station verify the effectiveness of the proposed method. Comparison experiments with convolutional neural network (CNN) and long short-term memory network (LSTM) shows that ConvLSTM can combine the advantages of both CNN and LSTM and significantly outperform CNN and LSTM in terms of forecast accuracy. Through further comparison with other five conventional methods including multilayer perceptron (MLP), support vector regression (SVR), extreme learning machine (ELM), classification and regression tree (CART) and gradient boosting decision tree (GBDT), ConvLSTM can significantly improve the forecast accuracy by more than 20% for most of the five methods and the superiorities of ConvLSTM are further verified.

## 1. Introduction

Nowadays, conventional fossil fuels are running out. Meanwhile, the exhaustive use of fossil fuels has caused serious environmental pollution. Developing renewable energy is one of the most important ways to solve the environmental problem and energy crisis [1,2]. Solar energy is a clean, low-carbon and inexhaustible energy. As one of the important ways to utilize solar energy, photovoltaic (PV) power generation has been developing rapidly in the recent years [3,4]. PV power are affected by solar irradiation, temperature, cloud, weather etc[ 5 ]. These factors cause the frequent fluctuation of PV power, which is harmful to the electricity grid and restricts the further application of PV power [6]. Accurate PV power forecast can ensure the electricity grid stability and promote the large-scale utilization of PV power, and thus has great significance.

Many methods have been developed to forecast PV power. Currently, the PV power forecast methods can be divided into two main categories: physical modelling-based method and historical data-based method. Compared with physical modelling-based method, historical data-based method only relies on historical data and does not need much physical prior knowledge. Thus, historical data-based method is becoming one of the most popular PV power forecast methods.

---
* Corresponding author
E-mail address: mingliangbai@outlook.com (Mingliang Bai), 15B902036@hit.edu.cn (Xinyu Zhao), zhenhualong@hit.edu.cn (Zhenhua Long), jinfuliu@hit.edu.cn (Jinfu Liu), yudaren@hit.edu.cn (Daren Yu)

Currently, many researchers have developed many historical data-based PV power forecast method. Many statistical methods and machine learning methods, such as autoregressive integrated moving average (ARIMA) model [7], rough set [8-10], random forest [11], support vector regression (SVR) [12], extreme learning machine (ELM) [13,14], artificial neural network (ANN) [15] etc., have been widely used in PV power forecast. Atique et al. [16] used ARIMA to forecast the total daily solar energy generation. Li et al. [17] used autoregressive moving average with exogenous inputs (ARMAX) to forecast the power output of a grid connected photovoltaic system and achieved better performances than ARMA model. ARIMA and ARMAX can characterize the linear time-delay relationships in the time series data well. However, the time-delay relationships in actual PV power data are not linear due to various uncertain factors in reality. Thus, many nonlinear regression methods including random forest, SVR, ELM, ANN etc., are introduced in PV power forecast. Niu et al. [18] used random forecast and wavelet decomposition for PV power forecast. Liu et al.[19] used random forest ensemble of SVM for PV power forecast and achieved better performance than single SVM. Behera et al. [20] proposed an improved extreme learning machine algorithm for PV power forecast. Yang et al. [21] and Mei et al. [22] introduced rough set theory into PV power forecast. Lin et al. [23] used K-means and Elman recurrent neural network for PV power forecast. Cheng et al. [24] used GRNN with temperature and humidity weighted modification for PV power forecast and achieved good performance.

Although above statistical methods and machine learning methods have been widely used in PV power forecast, the forecast accuracy of these conventional methods can be significantly improved by deep learning. As the recent development of ANN as well as the major breakthrough in machine learning, deep learning [25] is now enjoying a boom. Deep learning has made tremendous achievements and won great success in many fields such as computer vision [26], natural language processing[27], autonomous vehicles [28], recommendation systems [29], finance market [30] etc. Many researchers have introduced deep learning into PV power forecast and significantly improved the forecast performance. Currently, deep belief network (DBN)[31], convolutional neural network (CNN) and long short-term memory (LSTM) network [32,33] are three most commonly used deep learning methods in PV power forecast. Chang et al. [34] compared DBN with four conventional methods including ARIMIA, back propagation neural network (BPNN), radial basis function neural network (RBFNN) and SVR, and reported that DBN significantly outperforms the four conventional methods. Zang et al. [35] compared CNN with random forecast, Gaussian process regression, BPNN and SVR, and reported that CNN obtained the optimal forecast performance. Abdel-Nasser et al. [36] compared LSTM network with three conventional methods including multiple linear regression (MLR), bagged regression trees (BRT) and neural networks and verified the superiorities of LSTM network. Mellit et al. [37] compared LSTM network with two conventional neural networks including Elman recurrent neural networks and nonlinear autoregressive network, and reported that LSTM network obtained the optimal performance among these methods.

Recently, convolutional long short-term memory (ConvLSTM) network [38], a new deep learning method with the advantages of both CNN and LSTM, has been widely applied in various time series forecast tasks and achieved great success. Shi et al. [38] first proposed ConvLSTM network, applied ConvLSTM network for precipitation forecast and reported significant improvement than LSTM network. Çiçek and Gören[39]introduced ConvLSTM network into the smartphone power management, and reported ConvLSTM significantly outperforms CNN and LSTM in terms of forecast accuracy. Chen et al. [40] used ConvLSTM network for short-term traffic flow prediction and reported better forecast performance than LSTM network. Lin et al. [41] used ConvLSTM network with self-attention for both video prediction and traffic prediction, and significantly improved the forecast performance when compared with LSTM network. Lee et al. [42] proposed a ConvLSTM-based method for stock market index forecast, performed experiments in three indexes including S&P500, KOSPI200 and FTSE100, and obtained significant improvement than CNN and LSTM network.

Inspired by the great success that ConvLSTM network achieved in the above time series forecast fields, it is promising that ConvLSTM network can be introduced into PV power forecast and significantly improve the PV power forecast performance. Therefore, this paper introduced ConvLSTM network into PV power forecast field to improve the forecast performance for the first time. Meanwhile, kernel density estimation is used to post-process the point forecast results of ConvLSTM and obtain the probabilistic confidence interval forecast results of PV power.

The main contributions of this paper are summarized as follows:

(1) Convolutional long short-term memory (ConvLSTM) network is introduced into photovoltaic (PV) power forecast. To the best of our knowledge, this is the first time that ConvLSTM network has been used in PV power forecast.

(2) A series of detailed comparison experiments are made and verify the superiorities of ConvLSTM network in forecast accuracy. To the best of our knowledge, this is the first time that the superiorities of ConvLSTM network has been verified in PV power forecast. Through comparison with CNN, LSTM and five conventional forecast methods, the significant superiorities of ConvLSTM are systematically verified.

(3) Kernel density estimation (KDE) is used to post-process the forecast results of ConvLSTM network. To the best of our knowledge, this is the first time that ConvLSTM has been combined with KDE for PV power forecast. Through the combination of ConvLSTM network and KDE, both accurate point forecast and reliable probabilistic confidence interval forecast of PV power are realized. The accurate point forecast and reliable probabilistic confidence interval can greatly contribute to the dispatch of electricity grid.

The rest of this paper is organized as follows. Section 2 introduces ConvLSTM network and kernel density estimation and gives the detailed procedure of the proposed probabilistic PV power forecast method. Section 3 performs detailed experiments and compared the proposed method with various conventional methods to verify its superiorities. Section 4 concludes the paper and outlines the future research orientation.

## 2. Method

This section will introduce all the methods used in this paper. ConvLSTM network incorporates the advantages of both convolutional neural network (CNN) and long short-term memory (LSTM) network. Thus, this section first introduces CNN and LSTM briefly in Section 2.1 and Section 2.2 respectively, then describes ConvLSTM network in Section 2.3 to better clarify its superiorities. Next, kernel density estimation (KDE) is introduced for probabilistic forecast in Section 2.4. Finally, the detailed procedure of applying deep ConvLSTM network and KDE for probabilistic PV power forecast is illustrated in Section 2.5.

### 2.1. Convolutional neural network

CNN introduces local perception and weight sharing, uses the convolutional layers to replace the fully-connected layers in conventional neural networks and thus has better feature extraction ability than conventional neural networks. Many researchers have verified that CNN can achieve much better performances than conventional neural networks in various fields, such as image recognition [26], natural language processing [27], time series forecast [35] etc. The typical architecture of CNN is given in Fig. 1, where Conv denotes the convolutional layer.

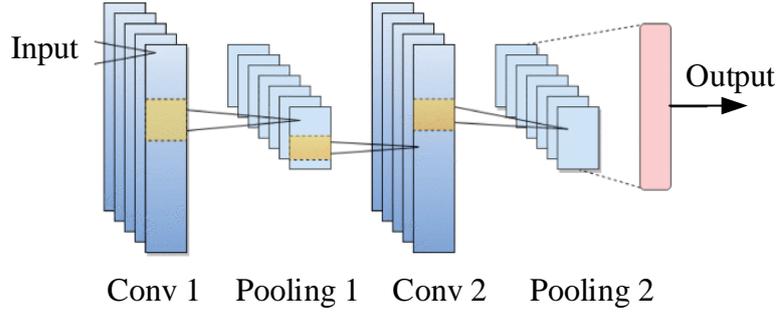

Conv 1    Pooling 1    Conv 2    Pooling 2

Fig. 1. Typical CNN architecture (adopted from literature [43])

In CNN, the $j$-th output feature map of the convolutional layer is expressed in Equation (1).

$$y^j = f\left(b^j + \sum_i w^{ij} * x^i\right) \quad (1)$$

where $x^i$ is the $i$-th input feature map, and $y^j$ is the $j$-th output feature map. $w^{ij}$ is the convolutional kernel, and $b^j$ is the bias term. The operation $*$ denotes convolution operation shown in Fig. 2. $f(\cdot)$ is the nonlinear activation function. CNN usually uses Rectified Linear Unit (ReLU) i.e., $f(x) = \max(0, x)$, as the activation function.

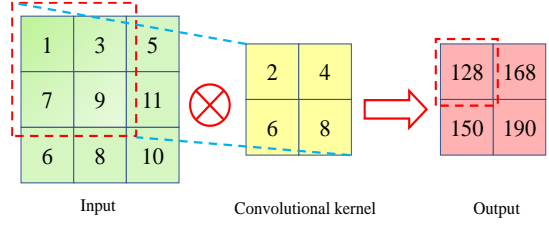

Fig. 2. Convolution operation

## 2.2. Long short-term memory network

Long short-term memory (LSTM) network has achieved great success in sequence modelling and time series prediction. LSTM network can extract temporal information automatically and characterize the long-term dependency well through three gates: forget gate, input gate and output gate. Forget gate determines what historical information is preserved. Input gate determines what current information is finally inputted. Output gate determines what information is used as the final output. Fig. 3 shows the architecture of LSTM network. The detailed principle of LSTM is illustrated in Equation (2).

$$\begin{aligned} i_t &= \sigma\left(W_{xi}x_t + W_{hf}h_{t-1} + b_i\right) \\ f_t &= \sigma\left(W_{xf}x_t + W_{hf}h_{t-1} + b_f\right) \\ C_t &= f_t \circ C_{t-1} + i_t \circ \tanh\left(W_{xc}x_t + W_{hc}h_{t-1} + b_c\right) \\ o_t &= \sigma\left(W_{xo}x_t + W_{ho}h_{t-1} + b_o\right) \\ h_t &= o_t \circ \tanh(C_t) \end{aligned} \quad (2)$$

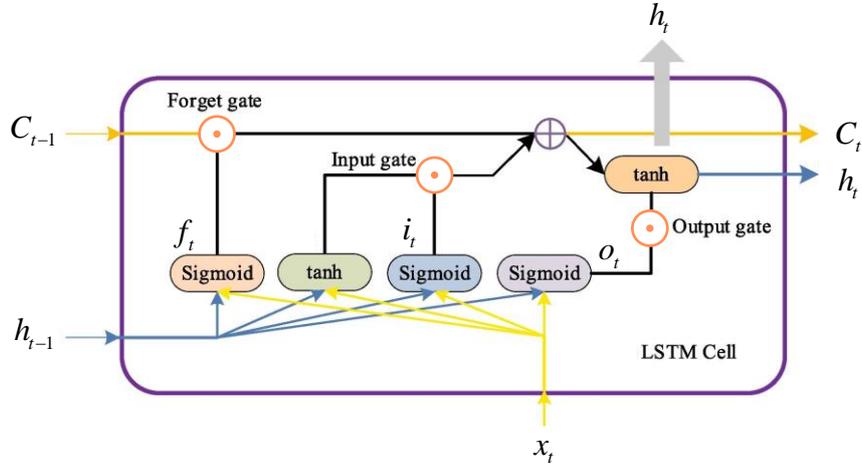

Fig. 3 LSTM network (adopted from literature [44]).

In Equation (2), $W_{xi}$, $W_{hi}$, $W_{xf}$, $W_{hf}$, $W_{xc}$, $W_{hc}$, $W_{xo}$ and $W_{ho}$ denote the weight term. $b_f$, $b_i$, $b_c$ and $b_o$ denote the bias term. The operation $\circ$ denotes element-wise product, $x_t$ denotes the current inputs and $h_{t-1}$ denotes the output of LSTM unit at the previous moment. $\sigma(.)$ is the Sigmoid nonlinear activation function defined in Equation (3) and $\tanh(.)$ is also a nonlinear activation function defined in Equation (4).

$$\sigma(x) = \frac{1}{1+e^{-x}} \quad (3)$$

$$\tanh(x) = \frac{e^x - e^{-x}}{e^x + e^{-x}} \tag{4}$$

## 2.3. Deep ConvLSTM network

Although CNN and LSTM have achieved great success in time series forecast, the forecast accuracy can be further improved by integrating the advantages of both CNN and LSTM. Shi et al. [38] introduced convolution operation into LSTM network, incorporated the advantages of both CNN and LSTM, proposed convolutional long short-term term memory (ConvLSTM) network and verified the superiorities of ConvLSTM network in precipitation forecast. Various literatures have verified the superiorities of ConvLSTM in time series forecast tasks including traffic flow forecast [40], video forecast [41], stock market forecast [42] etc. Thus, this paper introduces ConvLSTM network into PV power forecast. The typical architecture of ConvLSTM is shown in Fig. 4 and its detailed mathematical principle is illustrated in Equation (5).

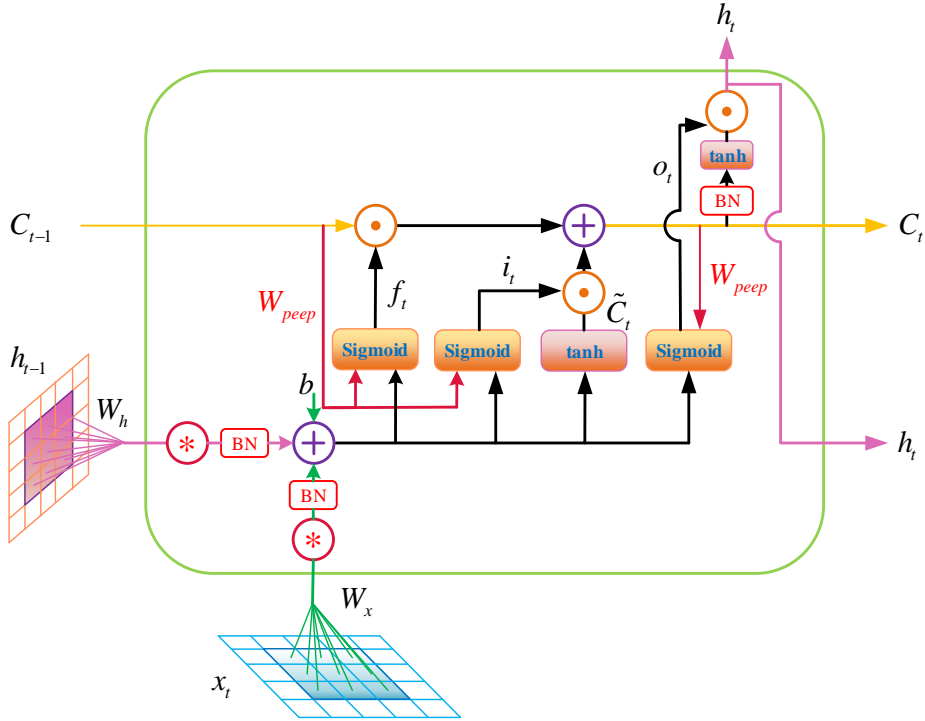

Fig. 4 ConvLSTM network

$$\begin{aligned}
i_t &= \sigma\left(W_{xi} * x_t + W_{hi} * h_{t-1} + W_{ci} \circ C_{t-1} + b_i\right) \\
f_t &= \sigma\left(W_{xf} * x_t + W_{hf} * h_{t-1} + W_{cf} \circ C_{t-1} + b_f\right) \\
C_t &= f_t \circ C_{t-1} + i_t \circ \tanh\left(W_{xc} * x_t + W_{hc} * h_{t-1} + b_c\right) \\
o_t &= \sigma\left(W_{xo} * x_t + W_{ho} * h_{t-1} + W_{co} \circ C_t + b_o\right) \\
h_t &= o_t \circ \tanh\left(C_t\right)
\end{aligned} \tag{5}$$

There are many similarities between Equation (5) and Equation (2). Like LSTM, $b_f$, $b_i$, $b_c$ and $b_o$ in Equation (5) also denote the bias term, the operation $\circ$ in Equation (5) also denotes element-wise product, $x_t$ in Equation (5) also denotes the current inputs, $h_{t-1}$ in Equation (5) also denotes the output of LSTM unit at the previous moment, and $\sigma(.)$ as well as $\tanh(.)$ in Equation (5) are also nonlinear activation functions defined in Equation (3) and Equation (4) respectively. Compared with LSTM, ConvLSTM introduces the convolution operation $*$ in CNN (see Fig. 2) into LSTM and replace the conventional weight term with the convolutional kernel. Thus, $W_{xi}$, $W_{hi}$, $W_{xf}$, $W_{hf}$, $W_{xc}$, $W_{hc}$, $W_{xo}$ and $W_{ho}$ in Equation (5) denote the convolutional kernel in Equation (5), which is the main improvements of ConvLSTM. Besides, ConvLSTM also introduces the peep weights ($W_{ci}$, $W_{cf}$ and $W_{co}$ in Equation (5)) into conventional LSTM to improve the ability to model long-range dependencies [38]. Through these improvements, ConvLSTM obtains significantly better performance than CNN and LSTM.

When applying ConvLSTM network to PV power forecast, many ConvLSTM layers are stacked together and thus the deep ConvLSTM network shown in Fig. 5 forms. Through deep ConvLSTM network, the time-delayed dependencies in PV power data can be characterized well and accurate PV power forecast can be realized.

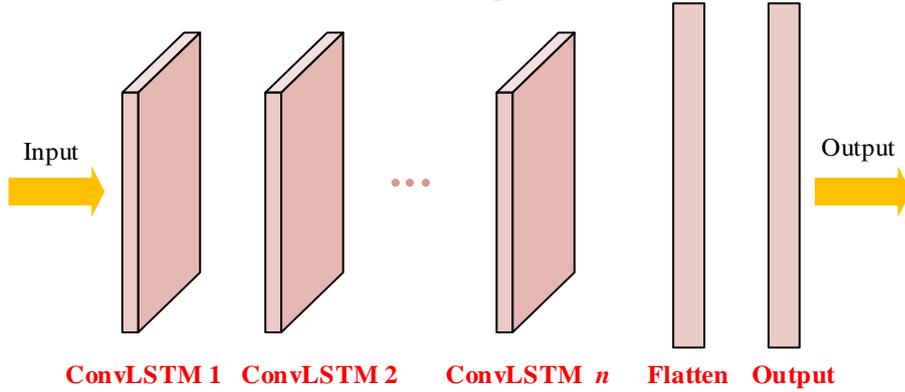

Fig. 5 Deep ConvLSTM network

### 2.4. Kernel density estimation

Although deep ConvLSTM network can achieve accurate forecast of PV power, it cannot give the probabilistic confidence interval. Kernel density estimation (KDE) is currently one of the most popular non-parameter methods to estimate the probabilistic density function (PDF) of data. Thus, KDE is used to post-process the forecast results of ConvLSTM network and give probabilistic confidence interval. Given a sequence $X$ with $N$ 1-dimensional observations $x_1, x_2, ..., x_N$, KDE can estimate PDF $f(\cdot)$ through Equation (6) [45]:

$$\hat{f}_1(x) = \frac{1}{Nh_1} \sum_{i=1}^{N} Ker(\frac{x - x_i}{h_1}) \qquad (6)$$

where $h_1$ denotes the bandwidth, and $Ker(\cdot)$ is the kernel function that can be Gaussian, Quadratic, or Epanechnikov etc[46,47]. It is proved that $\hat{f}_1(\cdot)$ can converge to $f(\cdot)$ efficiently if $N$ is large enough, and the accuracy of KDE isn't determined by the type of $Ker(\cdot)$ [48]. Therefore, this paper uses Gaussian kernel, which is a commonly used kernel function in KDE and can be formulated by Equation (7):

$$\text{Ker}(z) = e^{-z^2/2} / \sqrt{2\pi} \qquad (7)$$

When applying KDE to the d-dimensional space $X_i = (x_{i1}, x_{i2}, ..., x_{id})$, Equation (6) should be replaced by Equation (8):

$$\hat{f}_n(x) = \frac{1}{Nh_1 h_2 ... h_d} \sum_{i=1}^{N} \{\prod_{j=1}^{d} Ker_j(\frac{x_j - x_{ij}}{h_j})\} \qquad (8)$$

The optimum bandwidth value in Equation (8) can be selected by minimizing the mean integrated squared error. Through KDE, the joint PDF between the predicted PV power and the actual PV power can be estimated, the corresponding probabilistic confidence interval can be obtained and thus probabilistic interval forecast is realized.

## 2.5. Procedure of the proposed method

This paper applies ConvLSTM for short-term PV power forecast and uses KDE for probabilistic interval forecast of PV power. The main procedure of the proposed method is summarized in Fig. 6. The detailed descriptions of the procedure are as follows:

### 2.5.1. Data pre-processing

PV power measurements are almost zero at night and thus these measurements at night are deleted in the experiment. After that, the historical PV power data are divided into two parts in terms of time order: training set and test set. The first 80% is the training set and the rest 20% is the test set. The training set is used to train ConvLSTM network and the test set is used to evaluate the forecast performance.

### 2.5.2. Training

This process is implemented offline to train deep ConvLSTM network and obtain the estimated joint probabilistic density function (PDF). Two steps can be concluded.

● Step 1: train deep ConvLSTM network. Deep ConvLSTM network is trained using the training data. During the training process, the convolution kernels and bias terms in ConvLSTM are updated through error backpropagation. After training, the trained ConvLSTM network is obtained.

● Step 2: estimate joint PDF with KDE. The training inputs are first given to the trained deep ConvLSTM network to obtain the predicted PV power data. Then KDE is used to obtain the estimated joint PDF between the actual PV power data and the predicted PV power data.

### 2.5.3. Test

This process is implemented online for forecasting the future PV power. Current PV power measurements are inputted to the trained ConvLSTM network to obtain the predicted future PV power. Then the predicted future PV power is inputted to the estimated joint PDF to obtain the 95% confidence interval of the predicted future PV power, namely corresponding upper limits and lower limits.

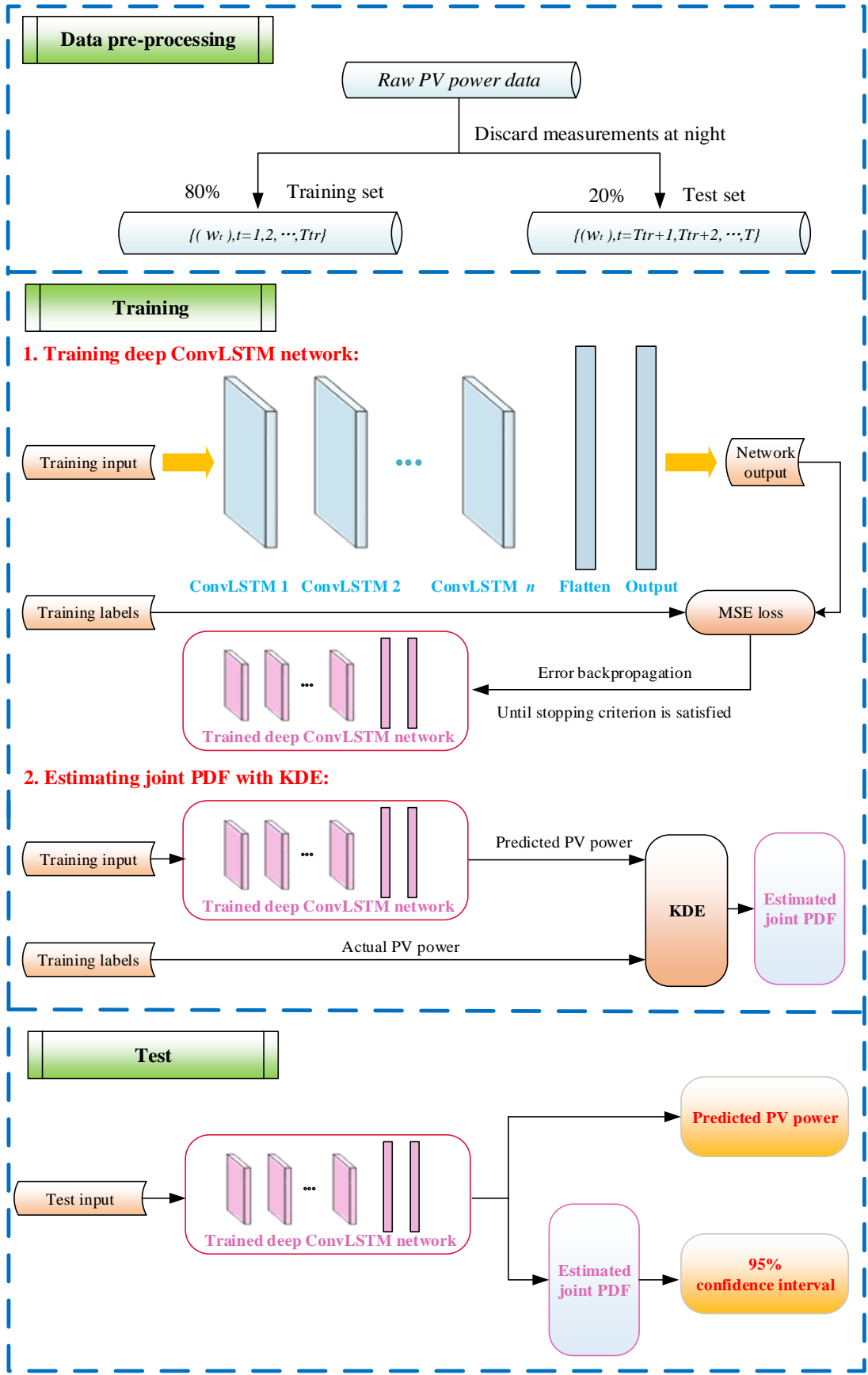

Fig. 6 Framework of the proposed method

## 3. Experiment
### 3.1. Data description

This paper uses actual measured PV power data in Belgium, which can be downloaded from the website [49], for forecast study. The monitored capacity of the studied PV power station is 2915.88 MW$_p$. The data are collected from January 1, 2015 to December 31, 2015 with a temporal resolution of 15 minutes, namely 96 datapoints per day. Fig. 7 shows the one-year PV power measurements used in this paper. This paper selects PV data from 5:00 a.m. to 8:00 p.m. per day for forecast study. The PV power measurements during the remaining period of a day are neglected because it is usually at night and PV power values are zero or nearly zero during this period. That is to say, each day have 60 samples after deleting samples at night.

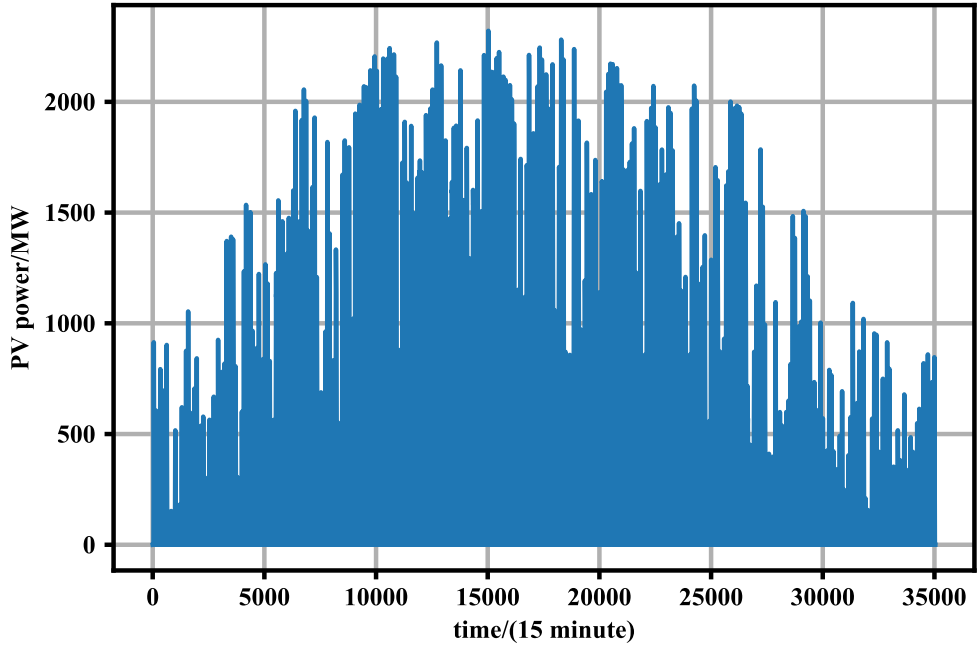

Fig. 7 Original PV power data

In the experiment, three metrics are used to evaluate the forecast performance. The first metric is mean absolute error (MAE) defined in Equation (9). The second metric is mean absolute relative error (MARE), MARE is scale-independent error metric and is usually defined by MAE normalized by a number. This paper refers to literature [31] and uses the installed capacity of PV power station $y_{rated}$ to normalize MAE. The definition of MARE is given in Equation (10). The third metric is root mean square error (RMSE) defined in Equation (11).

$$MAE = \frac{1}{n}\sum_{i=1}^{n}|y_i - \hat{y}_i| \qquad (9)$$

$$MARE = \frac{1}{n}\sum_{i=1}^{n}\frac{|y_i - \hat{y}_i|}{y_{rated}} \times 100\% \qquad (10)$$

$$RMSE = \sqrt{\frac{1}{n}\sum_{i=1}^{n}(y_i - \hat{y}_i)^2} \qquad (11)$$

where $n$ is the number of samples, $y_i$ is the actual PV power and $\hat{y}_i$ is the predicted PV power.
### 3.2. Forecasting PV power with deep ConvLSTM network

This section uses deep ConvLSTM network for PV power forecast. ConvLSTM network is implemented by Keras library of Python programming language. PV power data are periodical and the period is one day. In the experimental data, each day has 96 samples and preserves 60 samples after deleting samples at night. Thus, this paper uses the previous 60 samples to predict the future one-step-ahead PV power, namely PV power in the future 15 minute. Then rolling prediction is used to realize four-step-ahead PV power forecast, namely one-hour-ahead forecast.

Deep ConvLSTM network may contain one ConvLSTM layers. When applying deep ConvLSTM network, this paper first performs experiments to evaluate the forecast performances in test sets. The change of corresponding MAE and RMSE with the number of ConvLSTM layers is shown in Fig. 8.

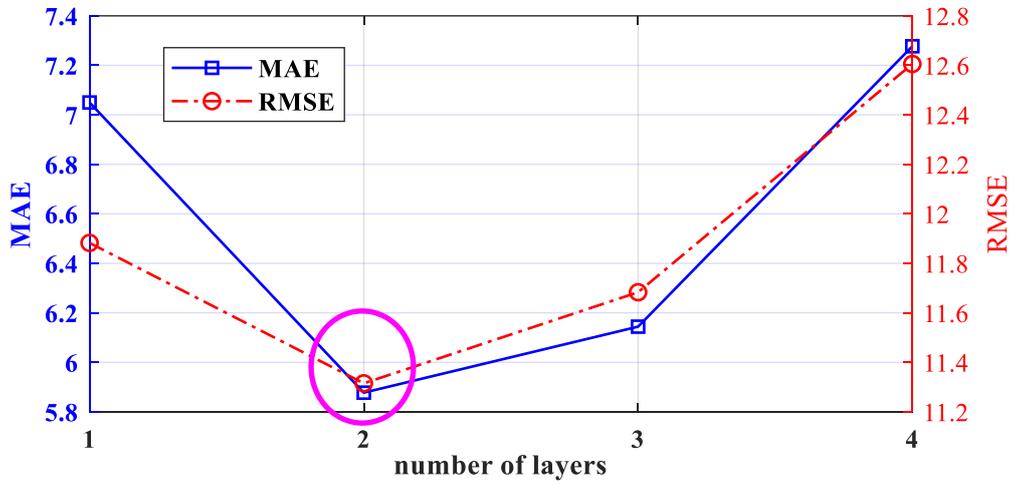

Fig. 8 Forecast performance versus the number of ConvLSTM layers

In Fig. 8, as the number of ConvLSTM layers increase, MAE and RMSE first decrease and then increase. When the number of ConvLSTM layers is 2, the minimal MAE and RMSE can be obtained. Thus, this paper selects two ConvLSTM layers in the experiments. Corresponding forecast performance is given in Table 1. Comparison between the actual PV power and the predicted PV power for one-step-ahead, two-step-ahead, three-step-ahead and four-step-ahead prediction is shown in Fig. 9-Fig. 12. Note that only part of test data is shown in Fig. 9-Fig. 12 for clear demonstration of results.

Table 1 Forecasting performance of ConvLSTM

|      | 1step   | 2step   | 3step   | 4step   |
| ---- | ------- | ------- | ------- | ------- |
| MAE  | 5.8775  | 12.1508 | 19.0380 | 26.0495 |
| RMSE | 11.3152 | 22.3553 | 34.7191 | 47.9319 |
| MARE | 0.20%   | 0.42%   | 0.65%   | 0.89%   |

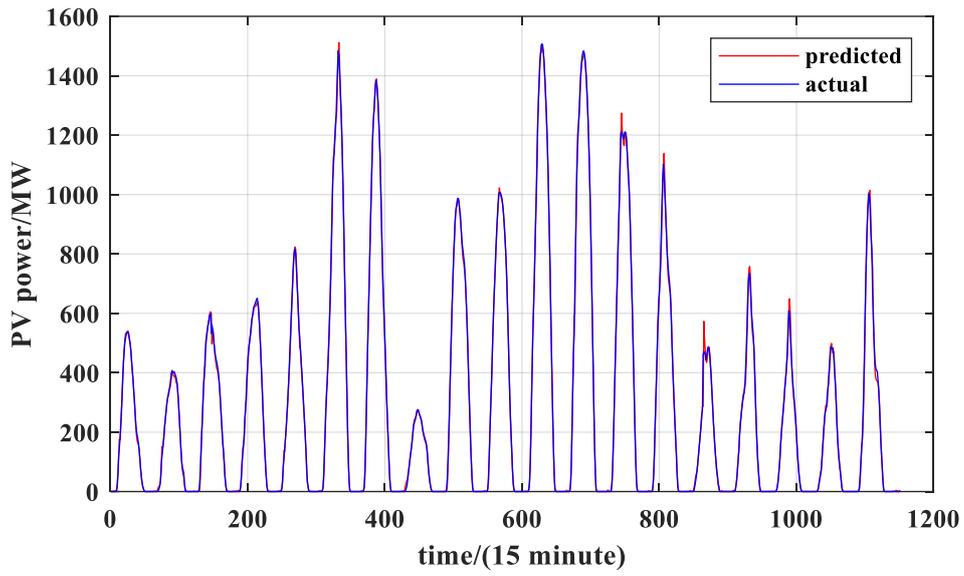

Fig. 9 One-step PV power prediction result of ConvLSTM

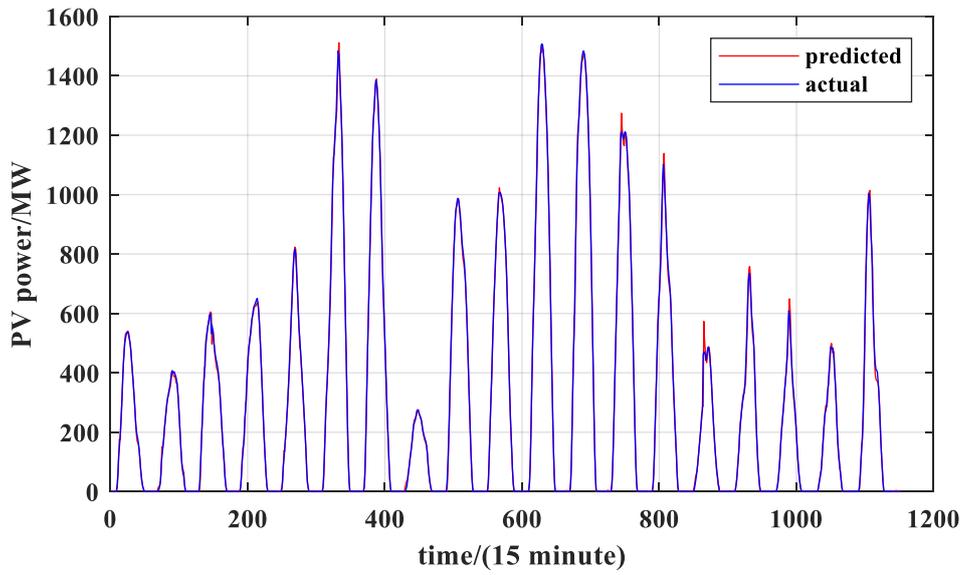

Fig. 10 Two-step PV power prediction result of ConvLSTM

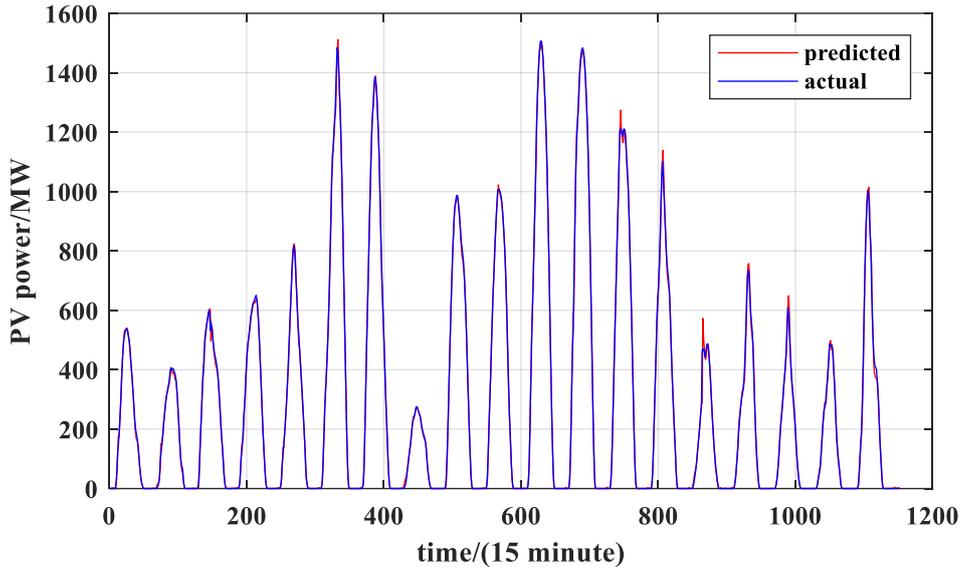

Fig. 11 Three-step PV power prediction result of ConvLSTM

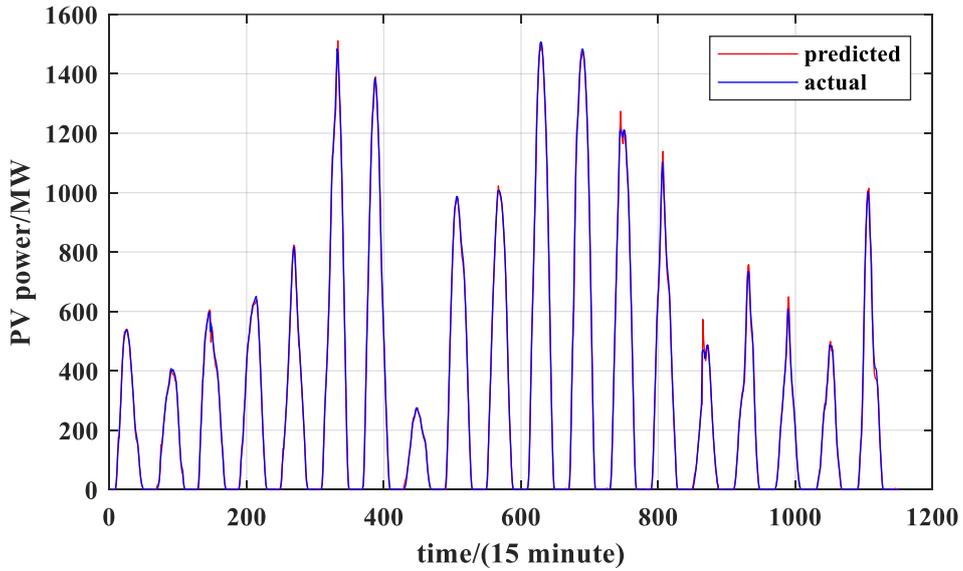

Fig. 12 Four-step PV power prediction result of ConvLSTM

From Table 1 and Fig. 9-Fig. 12, it is observed that the predicted values are close to the actual values and that the prediction errors are small. To measure how the predicted PV power is close to the actual PV power, this paper draws figures with the predicted PV power being the x-axis and the actual PV power being the y-axis, relevant figures are shown in Fig. 13. Besides, Pearson correlation coefficient $\rho$ defined in Equation (12) and coefficient of determination $R^2$ defined in Equation (13) are used to measure how the predicted values are close to the actual values quantitatively. For Pearson correlation coefficient $\rho$, it is within the interval $[-1,1]$, the value $-1$ means totally negative correlation and the value $1$ means totally positive correlation. Larger absolute value of Pearson correlation coefficient, namely $|\rho|$, means strong correlation relationship.

For coefficient of determination $R^2$, it measures the fitting performance and is less than or equal 1. The closer to 1 $R^2$ is, the better fitting performance the algorithms obtain. The results of Pearson correlation coefficient and coefficient of determination for deep ConvLSTM network are given in Table 2.

$$\rho = \frac{\sum_{i=1}^{n}\left[\left(y_i - \frac{1}{n}\sum_{t=1}^{n} y_i\right)\left(\hat{y}_i - \frac{1}{n}\sum_{i=1}^{n}\hat{y}_i\right)\right]}{\sqrt{\sum_{i=1}^{n}\left(y_i - \frac{1}{n}\sum_{i=1}^{n} y_i\right)^2 \sum_{i=1}^{n}\left(\hat{y}_i - \frac{1}{n}\sum_{i=1}^{n}\hat{y}_i\right)^2}} \quad (12)$$

$$R^2 = 1 - \frac{\sum_{i=1}^{n}(y_i - \hat{y}_i)^2}{\sum_{i=1}^{n}\left(y_i - \frac{1}{n}\sum_{i=1}^{n} y_i^2\right)^2} \quad (13)$$

where $n$ is the number of samples, $y_i$ is the actual PV power and $\hat{y}_i$ is the predicted PV power.

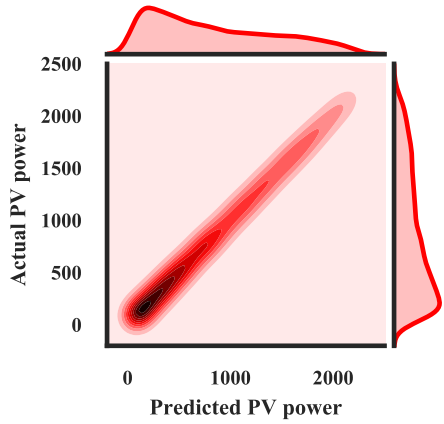

(a) 1-step training result

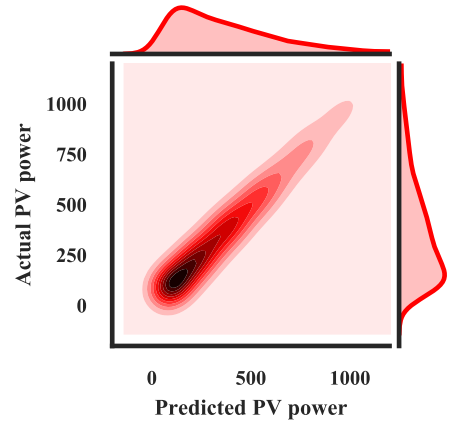

(b) 1-step test result

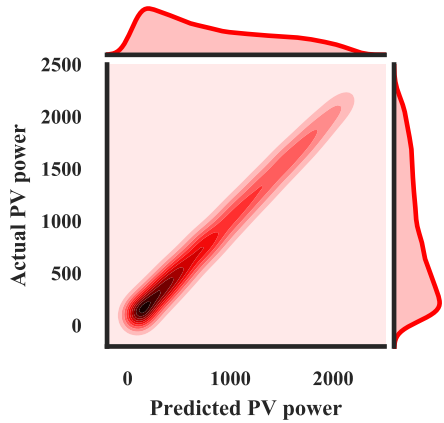

(c) 2-step training result

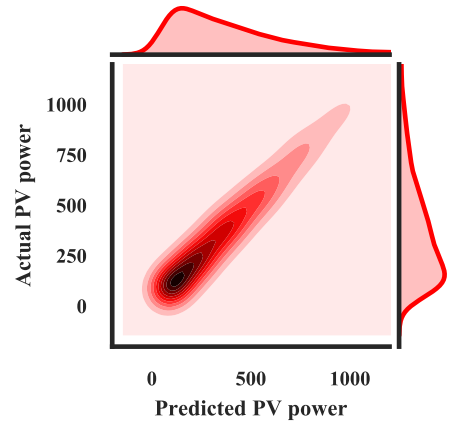

(d) 2-step test result

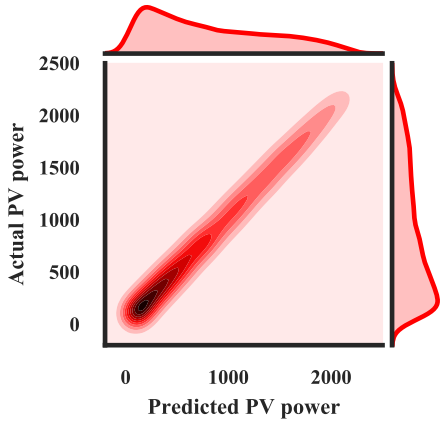
(e) 3-step training result

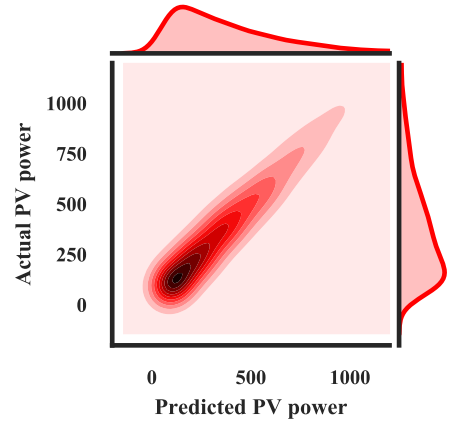
(f) 3-step test result

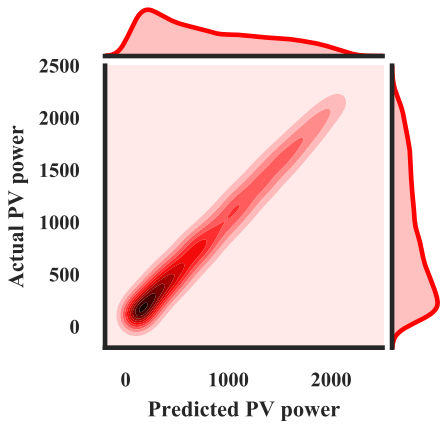
(g) 4-step test result

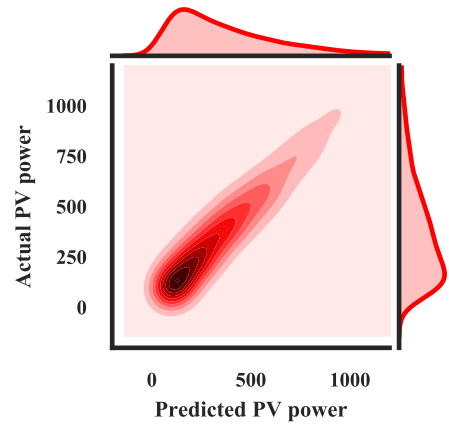
(h) 4-step test result

Fig. 13 Correlation between the predicted PV power and actual PV power

Table 2 $R^2$ and Pearson correlation of ConvLSTM forecast results

|  | 1step | 2step | 3step | 4step |
| --- | --- | --- | --- | --- |
| Pearson correlation | 0.9992 | 0.9970 | 0.9928 | 0.9864 |
| $R^2$ | 0.9984 | 0.9938 | 0.9851 | 0.9715 |

Table 2 and Fig. 13 show that the predicted PV power is close to the actual PV power. $R^2$ and Pearson correlation coefficient are both more than 0.97, which is very close to 1. This shows that there is strong positive correlation relationship between the predicted values and the actual values and that the fitting performance is satisfying. This further verifies the effectiveness of deep ConvLSTM network in PV power forecast.

### 3.3. Comparison with other methods

#### 3.3.1. Comparison with CNN and LSTM

ConvLSTM combines the advantages of both CNN and LSTM. To verify the superiorities of ConvLSTM, it is compared with CNN and LSTM. In the experiment, the hyperparameters of CNN and LSTM are the same as ConvLSTM, and CNN and LSTM both uses two CNN layers or two LSTM layers. Corresponding forecast performance comparison is shown in Table 3.

From Table 3, it is observed that ConvLSTM network obtains much smaller forecast errors than CNN and LSTM. This shows that ConvLSTM network significantly outperforms CNN and LSTM in PV power forecast.

Table 3 Comparison among ConvLSTM, CNN and LSTM

|  |  | 1step | 2step | 3step | 4step |
|---|---|---|---|---|---|
| MAE | **ConvLSTM** | **5.8775** | **12.1508** | **19.0380** | **26.0495** |
|  | CNN | 7.4850 | 16.0763 | 24.9516 | 33.6114 |
|  | LSTM | 7.5654 | 15.8668 | 25.5033 | 36.0215 |
| RMSE | **ConvLSTM** | **11.3152** | **22.3553** | **34.7191** | **47.9319** |
|  | CNN | 12.1965 | 24.5621 | 38.5813 | 53.6948 |
|  | LSTM | 13.6949 | 26.9390 | 41.5859 | 56.9543 |
| MARE | **ConvLSTM** | **0.20%** | **0.42%** | **0.65%** | **0.89%** |
|  | CNN | 0.26% | 0.55% | 0.86% | 1.15% |
|  | LSTM | 0.26% | 0.54% | 0.87% | 1.24% |
| Pearson correlation | **ConvLSTM** | **0.9992** | **0.9970** | **0.9928** | **0.9864** |
|  | CNN | 0.9992 | 0.9968 | 0.9920 | 0.9845 |
|  | LSTM | 0.9988 | 0.9955 | 0.9893 | 0.9798 |
| $R^2$ | **ConvLSTM** | **0.9984** | **0.9938** | **0.9851** | **0.9715** |
|  | CNN | 0.9982 | 0.9925 | 0.9815 | 0.9643 |
|  | LSTM | 0.9977 | 0.9910 | 0.9786 | 0.9598 |

### 3.3.2. Comparison with other conventional forecast methods

After comparing deep ConvLSTM network with CNN and LSTM, this section compares it with five conventional forecast methods to further verify its superiorities in PV power forecast. The compared methods include multilayer perceptron (MLP) [50], support vector regression (SVR) [51], extreme learning machine (ELM) [52], classification and regression tree (CART) [53] and gradient boosting decision tree (GBDT) [54]. MLP has an input layer, a hidden layer and an output layer, and MLP is trained through error backpropagation. ELM also has an input layer, a hidden layer and an output layer. But ELM is trained by computing Moore-Penrose generalized inverse matrix rather than iterative error backpropagation, and thus it is much faster than MLP during the training process. SVR uses kernel method to map the original data to a high dimensional space, so that an approximately linear regression can be used for regression in this space. Radial basis function (RBF) kernel is the most common kernel function. CART performs regression through splitting a series of branches in a tree. GBDT is a boosting ensemble of trees. All the five compared methods are commonly used in various forecast tasks.

In the experiment, among the five compared methods, MLP is implemented by Keras library of Python programming language, and the rest four methods are implemented by scikit-learn library [55,56] of Python programming language. The comparison results are listed in Table 4-Table 8.
.

Table 4 MAE comparison between ConvLSTM and other conventional methods

|  | 1step | 2step | 3step | 4step |
|---|---|---|---|---|
| **ConvLSTM** | **5.8775** | **12.1508** | **19.0380** | **26.0495** |
| MLP | 8.1593 | 16.8582 | 26.6780 | 36.9935 |
| SVR | 11.6430 | 23.4137 | 36.6355 | 51.6411 |
| ELM | 7.9055 | 17.7113 | 30.1119 | 44.6491 |
| CART | 11.6283 | 20.6052 | 29.7346 | 38.4491 |
| GBDT | 10.6660 | 18.9143 | 26.9459 | 34.7202 |

Table 5 RMSE comparison between ConvLSTM and other conventional methods

|  | 1step | 2step | 3step | 4step |
|---|---|---|---|---|
| ConvLSTM | **11.3152** | **22.3553** | **34.7191** | **47.9319** |
| MLP | 15.2727 | 31.0658 | 49.1673 | 68.2710 |
| SVR | 20.7915 | 41.7145 | 64.9068 | 90.4310 |
| ELM | 12.7495 | 26.6740 | 43.6049 | 62.9366 |
| CART | 22.5821 | 39.4859 | 56.3887 | 73.4205 |
| GBDT | 17.0850 | 31.2167 | 45.0208 | 58.5211 |

Table 6 MARE comparison between ConvLSTM and other conventional methods

|  | 1step | 2step | 3step | 4step |
|---|---|---|---|---|
| **ConvLSTM** | **0.20%** | **0.42%** | **0.65%** | **0.89%** |
| MLP | 0.28% | 0.58% | 0.91% | 1.27% |
| SVR | 0.40% | 0.80% | 1.26% | 1.77% |
| ELM | 0.27% | 0.61% | 1.03% | 1.53% |
| CART | 0.40% | 0.71% | 1.02% | 1.32% |
| GBDT | 0.37% | 0.65% | 0.92% | 1.19% |

Table 7 Pearson correlation comparison between ConvLSTM and other conventional methods

|  | 1step | 2step | 3step | 4step |
|---|---|---|---|---|
| **ConvLSTM** | **0.9992** | **0.9970** | **0.9928** | **0.9864** |
| MLP | 0.9990 | 0.9964 | 0.9915 | 0.9842 |
| SVR | 0.9974 | 0.9896 | 0.9751 | 0.9525 |
| ELM | 0.9991 | 0.9964 | 0.9908 | 0.9816 |
| CART | 0.9969 | 0.9907 | 0.9814 | 0.9688 |
| GBDT | 0.9982 | 0.9940 | 0.9876 | 0.9791 |

Table 8 $R^2$ comparison between ConvLSTM and other conventional methods

|  | 1step | 2step | 3step | 4step |
|---|---|---|---|---|
| **ConvLSTM** | **0.9984** | **0.9938** | **0.9851** | **0.9715** |
| MLP | 0.9971 | 0.9880 | 0.9700 | 0.9422 |
| SVR | 0.9946 | 0.9784 | 0.9477 | 0.8985 |
| ELM | 0.9980 | 0.9912 | 0.9764 | 0.9508 |
| CART | 0.9937 | 0.9807 | 0.9606 | 0.9332 |
| GBDT | 0.9964 | 0.9879 | 0.9749 | 0.9575 |

From Table 4-Table 8, it is observed that ConvLSTM obtains much smaller forecast errors than the other five methods. Thus, ConvLSTM significantly outperforms the compared five methods in PV power forecast.

### 3.3.3. Improvement percentage on forecast performance

To further show the superiorities of the proposed method, the improvement percentage of ConvLSTM is computed. For MAE and RMSE, the improvement percentage $\delta_{MAE}$ and $\delta_{RMSE}$ are defined by Equation (14) and Equation (15) respectively. $\delta_{MAE}$ and $\delta_{RMSE}$ are computed for all the compared methods in the experiments. Corresponding results are listed in Table 9, Table 10 and Fig. 14. Note that the improvement percentage of MARE is not listed. In this paper, MARE is MAE normalized by the installed capacity and the installed capacity is a fixed value. The improvement percentage of MARE obviously equals the improvement percentage of MAE $\delta_{MAE}$, and thus the improvement percentage of MARE is not listed in Table 9, Table 10 and Fig. 14.

$$\delta_{MAE} = \frac{MAE_{ConvLSTM} - MAE_{others}}{MAE_{others}} \times 100\% \qquad (14)$$

$$\delta_{RMSE} = \frac{RMSE_{ConvLSTM} - RMSE_{others}}{RMSE_{others}} \times 100\% \qquad (15)$$

where $MAE_{ConvLSTM}$ and $RMSE_{ConvLSTM}$ are MAE and RMSE of deep ConvLSTM network, and $MAE_{others}$ as well as $RMSE_{others}$ are MAE and RMSE of other compared methods.

Table 9 MAE improvement when compared with other methods

|       | 1step   | 2step   | 3step   | 4step   |
|-------|---------|---------|---------|---------|
| CNN   | 21.48%  | 24.42%  | 23.70%  | 22.50%  |
| LSTM  | 22.31%  | 23.42%  | 25.35%  | 27.68%  |
| MLP   | 27.97%  | 27.92%  | 28.64%  | 29.58%  |
| SVR   | **49.52%** | **48.10%** | **48.03%** | **49.56%** |
| ELM   | 25.65%  | 31.40%  | 36.78%  | 41.66%  |
| CART  | 49.46%  | 41.03%  | 35.97%  | 32.25%  |
| GBDT  | 44.89%  | 35.76%  | 29.35%  | 24.97%  |

Table 10 RMSE improvement when compared with other methods

|       | 1step   | 2step   | 3step   | 4step   |
|-------|---------|---------|---------|---------|
| CNN   | 7.23%   | 8.98%   | 10.01%  | 10.73%  |
| LSTM  | 17.38%  | 17.02%  | 16.51%  | 15.84%  |
| MLP   | 25.91%  | 28.04%  | 29.39%  | 29.79%  |
| SVR   | 45.58%  | **46.41%** | **46.51%** | **47.00%** |
| ELM   | 11.25%  | 16.19%  | 20.38%  | 23.84%  |
| CART  | **49.89%** | 43.38%  | 38.43%  | 34.72%  |
| GBDT  | 33.77%  | 28.39%  | 22.88%  | 18.09%  |

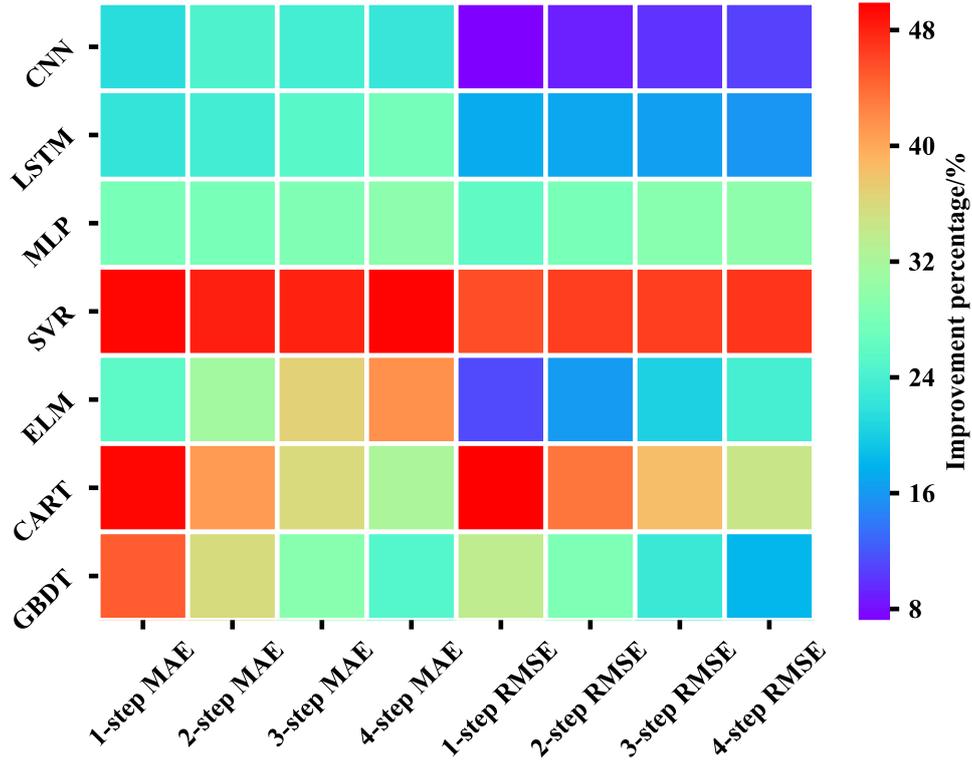

Fig. 14 Improvement percentage of ConvLSTM

From Table 9, Table 10 and Fig. 14, it is observed that ConvLSTM significantly improves the PV power forecast performance when compared with other methods. For MAE, ConvLSTM improves more than 20% for all the compared methods. For RMSE, ConvLSTM improves approximately 10% when compared with CNN, improves more than 15% when compared with LSTM, improves 20% or more when compared with most of the conventional methods including MLP, ELM, SVR, CART and GBDT. Thus, the proposed method significantly outperforms conventional methods in PV power forecast.

**3.4. Probabilistic PV power forecast**

After above experiments, the superiorities of ConvLSTM in PV power forecast are verified. This section uses KDE to post-process the ConvLSTM forecast data and realize probabilistic forecast. The predicted PV power data in training set and the actual PV power data in training set are given to KDE algorithm and the joint probabilistic density between the predicted PV power and the actual PV power is obtained. Fig. 15-Fig. 18 gives the joint probabilistic density estimation results for one-step, two-step, three-step and four-step prediction respectively in the form of pseudo-color maps. Through the estimated probabilistic density, the 95% confidence interval can be given after the predicted PV power data of test set are obtained through deep ConvLSTM network. The probabilistic forecast results in test sets are given in Fig. 19-Fig. 22.

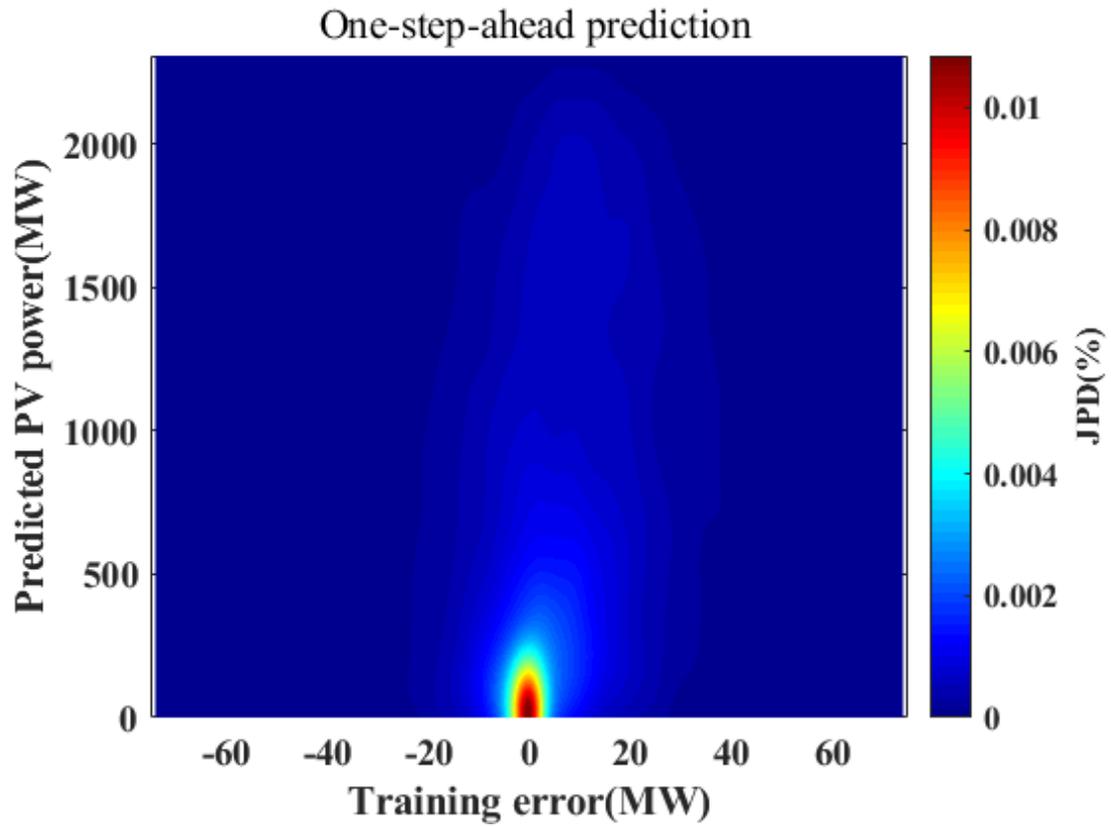

Fig. 15 Joint probabilistic distribution estimation for one-step prediction

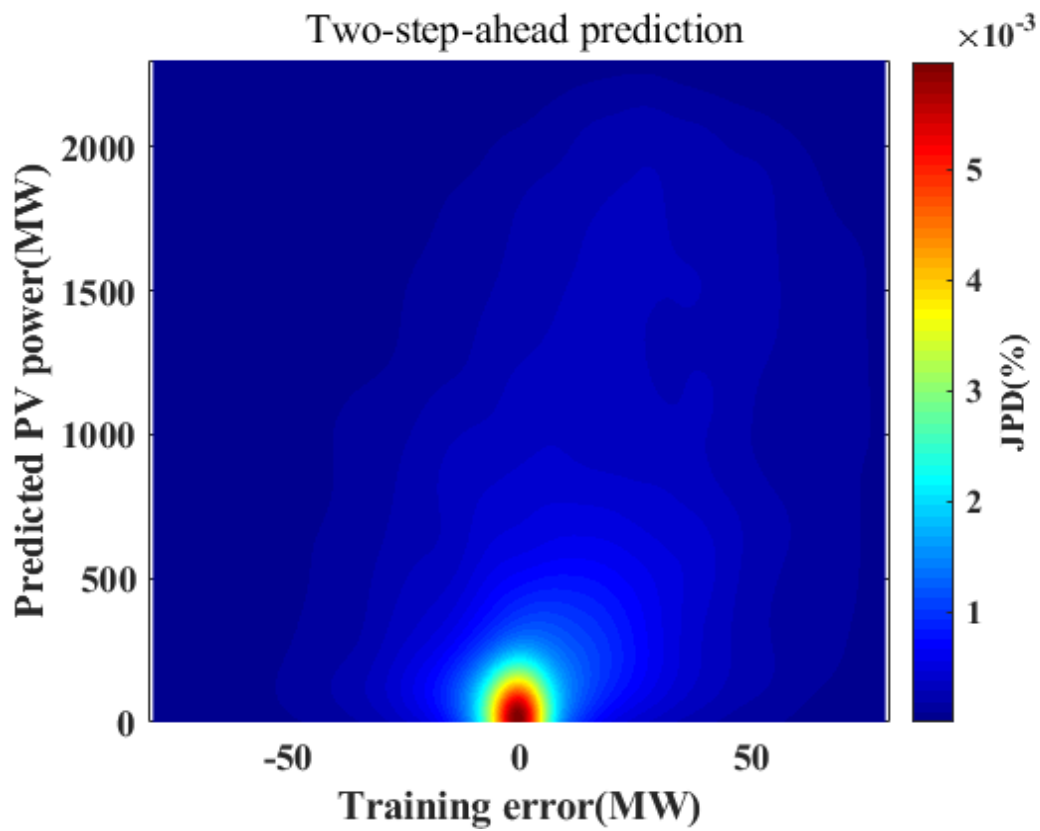

Fig. 16 Joint probabilistic distribution estimation for two-step prediction

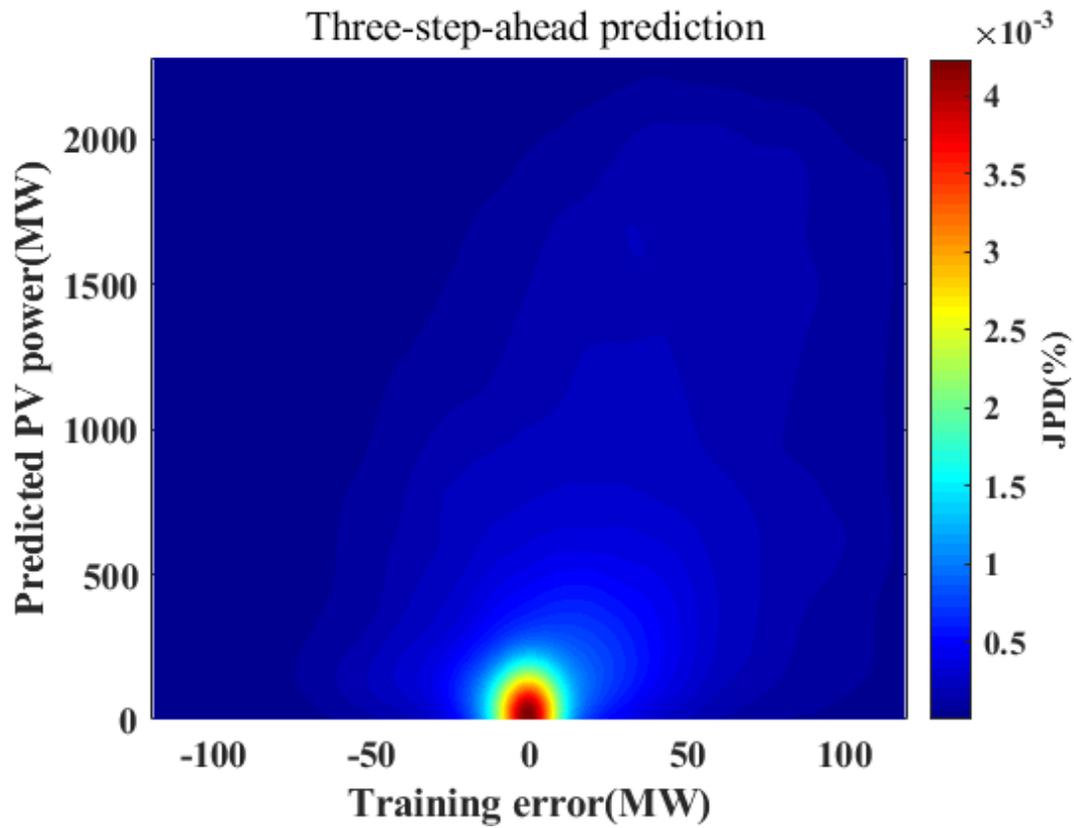

Fig. 17 Joint probabilistic distribution estimation for three-step prediction

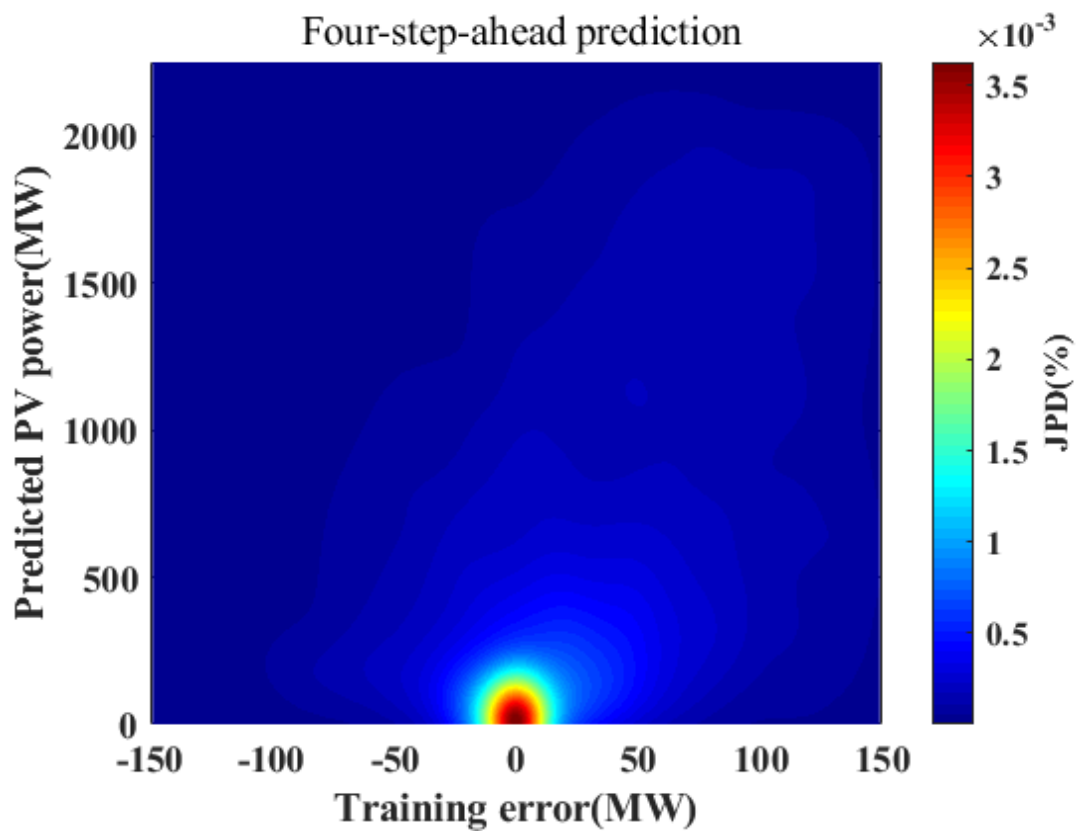

Fig. 18 Joint probabilistic distribution estimation for four-step prediction

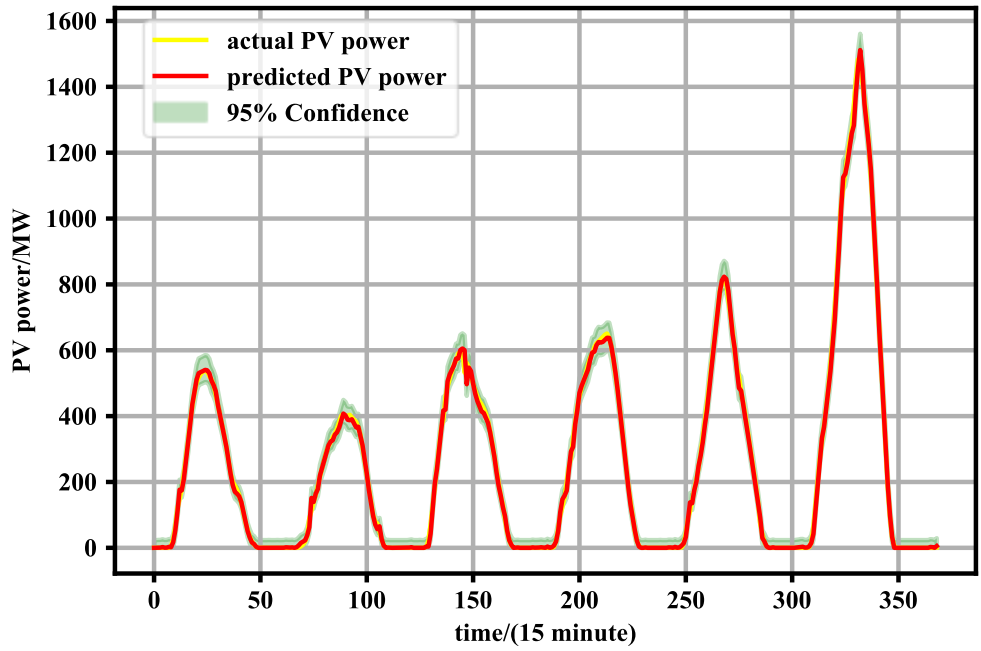

Fig. 19 One-step probabilistic interval forecast

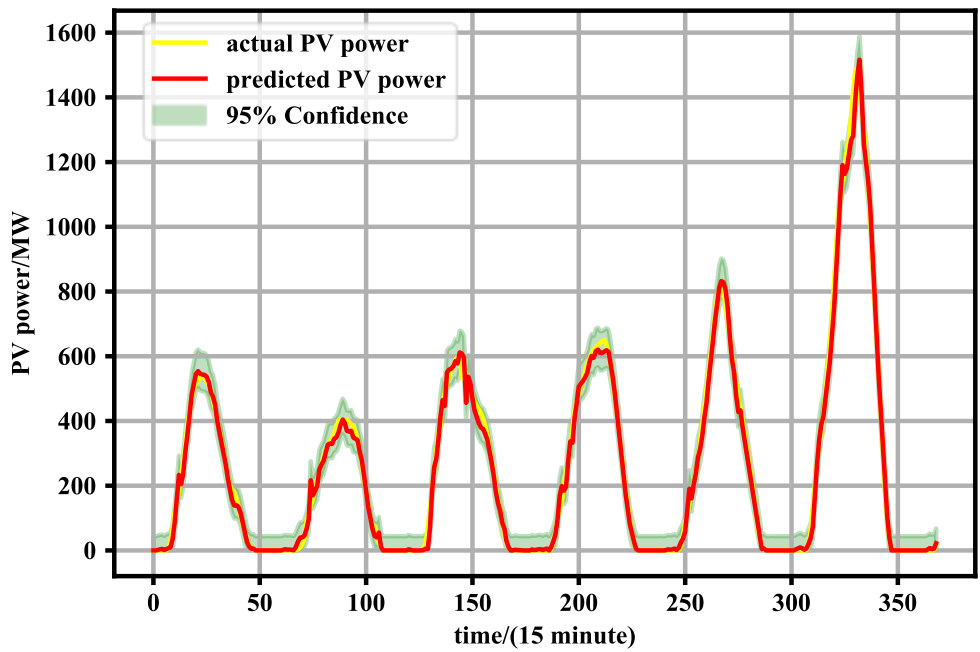

Fig. 20 Two-step probabilistic interval forecast

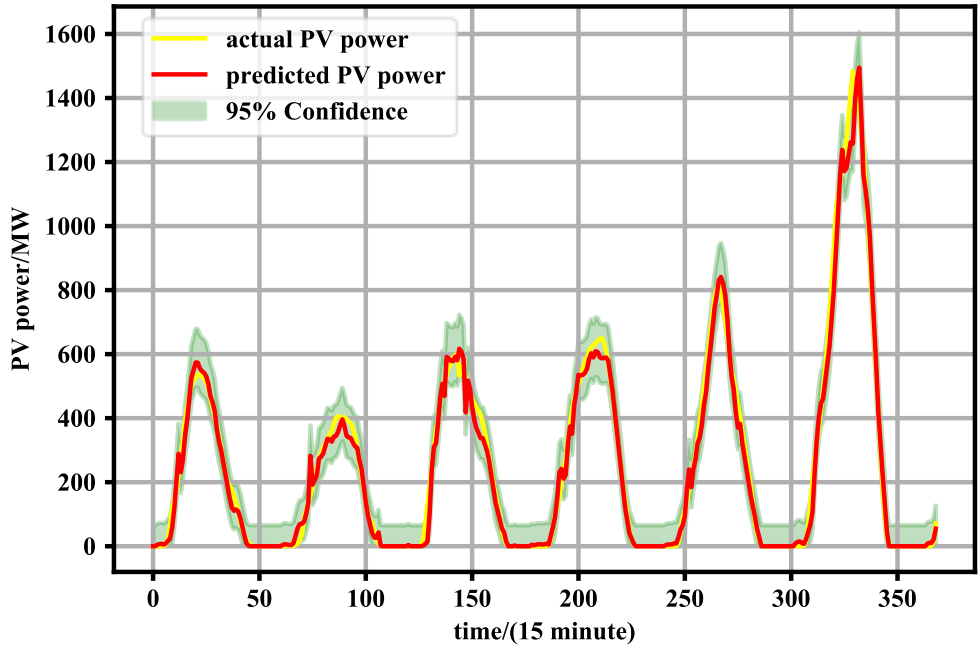

Fig. 21 Three-step probabilistic interval forecast

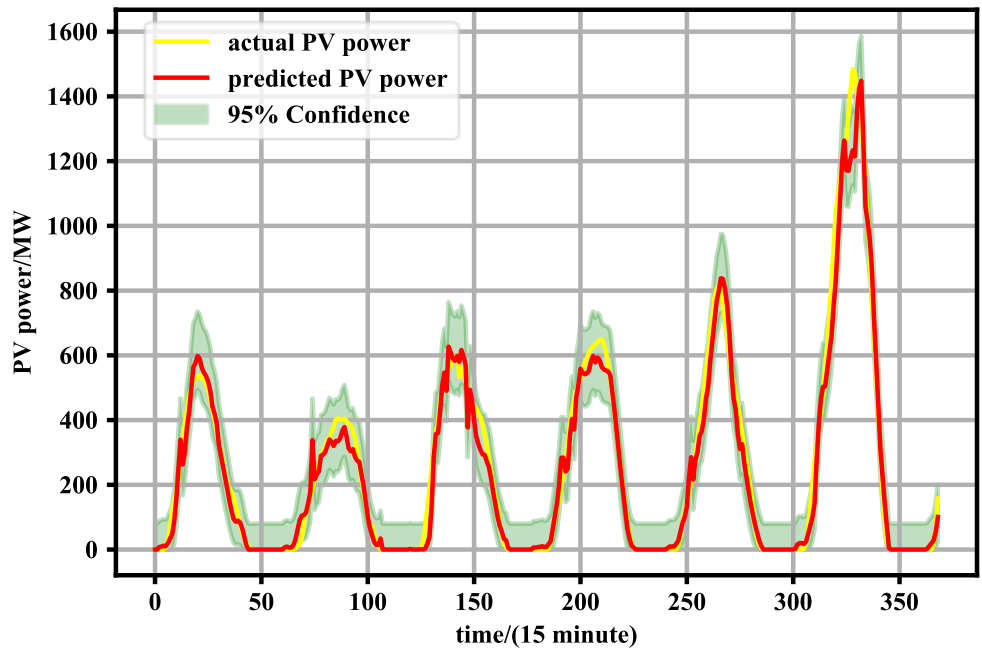

Fig. 22 Four-step probabilistic interval forecast

In Fig. 19-Fig. 22, the green shadows denote the 95% confidence interval, the red lines denote the actual PV power and the yellow lines denote the predicted PV power. It is observed that most of the actual PV power and the predicted PV power are within the given 95% confidence interval. This means that the obtained confidence interval is reliable and that the probabilistic forecast is effective.

To quantify the probabilistic PV power forecast performance of test set, two metrics, namely prediction interval coverage probability (PICP) and prediction interval average width (PIAW), are used. PICP represents the percentage of samples within the given confidence interval and PIAW is the average interval width. The two metrics are defined in Equation (16) and (17) respectively [45, 57]. PICP and PIAW of test PV power samples are listed in Table 11.

$$PICP = \frac{1}{n}\sum_{t=1}^{n} I_t \tag{16}$$

$$PIAW = \frac{1}{n}\sum_{t=1}^{n}(\hat{u}_{t,upper} - \hat{u}_{t,lower}) \tag{17}$$

where $n$ denotes the number of test samples, $y_t$ denotes the actual measured PV power, $\hat{u}_{t,upper}, \hat{u}_{t,lower}$ denote the upper and lower boundaries of forecast respectively, and $I_t$ is defined in Equation (18).

$$I_t = \begin{cases} 1, & \text{if } y_t \in [\hat{u}_{t,lower}, \hat{u}_{t,upper}] \\ 0, & \text{if } y_t \notin [\hat{u}_{t,lower}, \hat{u}_{t,upper}] \end{cases} \tag{18}$$

Table 11 Probabilistic forecast result

|  | 1-step | | 2-step | | 3-step | | 4-step | |
| --- | --- | --- | --- | --- | --- | --- | --- | --- |
|  | PICP | PIAW | PICP | PIAW | PICP | PIAW | PICP | PIAW |
| train | 0.9443 | 63.2817 | 0.9499 | 99.4890 | 0.9521 | 150.0820 | 0.9513 | 190.7061 |
| test | 0.9789 | 45.1331 | 0.9590 | 76.3159 | 0.9556 | 114.2159 | 0.9489 | 145.8368 |

From Table 11, it is observed that PICP approximates to 95% for both training data and test data. The goal of probabilistic forecast in this paper is to give 95% confidence interval. PICP means the percentage of samples within the given confidence interval. The fact the PICP approximates to 95% means that the obtained 95% probabilistic confidence interval is reliable enough. This verifies the effectiveness of the proposed probabilistic forecast method.

## 4. Conclusion and future work

Accurate photovoltaic power forecast is crucial to the large-scale application of photovoltaic power and the stability of electricity grid. To realize accurate photovoltaic power forecast, this paper introduces the deep convolutional long short-term memory (ConvLSTM) network into photovoltaic power forecast for the first time, and proposes a novel method for short-term probabilistic photovoltaic power forecast using deep convolutional long short-term memory (ConvLSTM) network and kernel density estimation (KDE). Through experiments in actual photovoltaic power data, the following conclusions can be drawn.

Firstly, this paper introduces a new forecast method, ConvLSTM network, into photovoltaic power for the first time. Through a series of forecast in a one-year measurement data of an actual photovoltaic power station in Belgium, the effectiveness of the proposed ConvLSTM-based forecast method is verified in detail. Experimental results show that the predicted values of ConvLSTM are close to the actual values and that the forecast errors are small. Pearson correlation analysis shows that there is strong positive correlation between the predicted values and the actual values, which can further verify the effectiveness of the proposed forecast method.

Secondly, this paper compares the photovoltaic power forecast performance of the proposed ConvLSTM-based method with convolutional neural network (CNN) and long short-term memory network (LSTM). Experimental results show that ConvLSTM can combine the advantages of both CNN and LSTM and significantly outperform CNN and LSTM in terms of forecast accuracy. To the best of our knowledge, this is the first time that the superiorities of ConvLSTM network over CNN and LSTM have been verified in photovoltaic power forecast.

Thirdly, this paper further compares the proposed method with other five conventional methods including multilayer perceptron (MLP), support vector regression (SVR), extreme learning machine (ELM), classification and regression tree (CART) and gradient boosting decision tree (GBDT). Experiments show that ConvLSTM can significantly improve the forecast accuracy by more than 20% for most of the five methods and the superiorities of ConvLSTM are further verified.

Fourthly, ConvLSTM network is combined with KDE to realize the probabilistic interval forecast. Through the combination of ConvLSTM and KDE, joint probabilistic density between the predicted values and actual values is estimated, and reliable 95% probabilistic confidence interval is given. Experiments shows that the percentage of samples within the given 95% confidence intervals approximates to 95% in both training set and test set. This proves that the proposed forecast method is effective and reliable.

In the future, we hope that ConvLSTM network and KDE can be used to realize accurate forecast and reliable probabilistic forecast of various energy parameters such as wind speed, wind power, load demand, electricity price and so on. Meanwhile, we also hope that ConvLSTM network can be used to the remaining useful life prediction of various industrial machines such as bearings, aeroengines, batteries and so on, to realize accurate prognostics and health management.

**Acknowledgement**

This work is supported by National Natural Science Foundation of China No. 51976042, National Science Technology Major Project of China No. 2017-I-0007-0008 and National Key R&D Program of China No. 2017YFB0902100. The authors would like to thank the anonymous reviewers for their valuable suggestions to refine this work.